\documentclass[preprint,3p,times,twocolumn,number]{elsarticle}
\usepackage{graphicx} 
\usepackage{lipsum}
\usepackage[dvipsnames]{xcolor}

\usepackage{todonotes}

\usepackage{ulem}
\usepackage{xspace}
\usepackage{tabularx}
\usepackage{booktabs}
\usepackage{array}
\usepackage{varwidth} 
\usepackage{makecell}
\usepackage{multirow}
\usepackage{url}
\usepackage[font=footnotesize,labelfont=bf]{subcaption}

\renewcommand\theadalign{bc}

\newcommand{\eg}{e.\,g.,\xspace}

\newcommand{\cf}{cf.\xspace}

\newcolumntype{C}{>{\centering\arraybackslash}X}

\begin{document}

\begin{frontmatter}



\title{Online Knowledge Integration for 3D Semantic Mapping: A Survey}


\author[dfki]{Felix Igelbrink} 
\author[dfki]{Marian Renz} 
\author[dfki]{Martin Günther} 
\author[uos_pbr]{Piper Powell} 
\author[dfki]{Lennart Niecksch} 
\author[dfki]{Oscar Lima} 
\author[dfki,uos_sis]{\mbox{Martin Atzmueller}} 
\author[dfki,uos_pbr]{Joachim Hertzberg} 

\affiliation[dfki]{organization={German Research Center for Artificial Intelligence\\ Research Department Plan-based Robot Control},
            city={Osnabrück},
            country={Germany}}
\affiliation[uos_pbr]{organization={Osnabrück University, Institute of Computer Science\\ Knowledge-Based Systems Group},
            city={Osnabrück},
            country={Germany}}
\affiliation[uos_sis]{organization={Osnabrück University, Institute of Computer Science\\ Semantic Information Systems Group},
            city={Osnabrück},
            country={Germany}}

\begin{abstract}

Semantic mapping is a key component of robots operating in and interacting with objects in structured environments. Traditionally, geometric and knowledge representations within a semantic map have only been
 loosely integrated. However, recent advances in deep learning now allow full integration of prior knowledge, represented as knowledge graphs or language concepts, into sensor data processing and semantic mapping pipelines. Semantic scene graphs and language models enable modern semantic mapping approaches to incorporate graph-based prior knowledge or to leverage the rich information in human language both during and after the mapping process. This has sparked substantial advances in semantic mapping, leading to previously impossible novel applications. This survey reviews these recent developments comprehensively, with a focus on online integration of knowledge into semantic mapping. We specifically focus on methods using semantic scene graphs for integrating symbolic prior knowledge and language models for respective capture of implicit common-sense knowledge and natural language concepts.

\end{abstract}


\end{frontmatter}



\section{Introduction}


For an autonomous mobile robot to perform its tasks, it must be able to localize itself in unfamiliar surroundings.
This requires a map of the environment, which is usually obtained via a
\textit{Simultaneous Localization and Mapping} (SLAM) approach  \cite{stachniss:etal:2016, Cadena2016SLAM}. These maps can typically be generated from sensor data in high quality and consider the geometric properties of the environment (\eg metric or topological in 2D or 3D), creating geometric maps with crucial information for localization, navigation, and obstacle avoidance. However, many real-life environments are not solely defined by their spatial properties and require a deeper understanding of the entities and structures the robot encounters.

Human-made environments in particular are highly dynamic and require specific interactions the robot must be prepared for, such as opening doors or manipulating objects. Additionally, many environments require a robot to follow certain, often unwritten rules to not behave unexpectedly or even dangerously, such as not blocking open doors, not moving unexpectedly around people, and not navigating into forbidden areas. Interacting with humans further requires a robot to be able to follow verbal commands from human operators, \eg ``Bring me my cup from the kitchen!'' Safe operation in such environments and fulfilling all required functions is intractable for a robot using only the spatial information of a geometric map, requiring solutions that provide the robot with further information to complete its tasks. 

One prominent approach to alleviate these problems is semantic mapping, \cf~\citet{nuchterSemanticMapsMobile2008b}:
\begin{quote}
    \textit{``A semantic map for a mobile robot is a map that contains, in addition to spatial information about the environment, assignments of mapped features to entities of known classes. Further knowledge about these entities, independent of the map contents, is available for reasoning in some knowledge base with an associated reasoning engine.''}
\end{quote}

Semantic maps thus add different kinds of information into a geometric map, such as sensor-derived information and external common-sense and specialized prior knowledge given by, \eg an external knowledge base. Such knowledge sources supply a robot with additional task-dependent information about its current environment and allow it to interpret and reason over abstract human concepts like rules and organization schemes, \eg where objects of a certain type are commonly expected.

Classically, the process of semantic mapping can be decomposed into multiple interconnected sub-problems \citep{hanSemanticMappingMobile2021}, which we review more extensively in Section~\ref{sec:nutshell}:
\begin{description}
    \item[1. Geometric mapping.] The information from a robot's 3D sensors is captured and aggregated into a suitable geometric representation.
    \item[2. Semantic information from sensor data.] Semantic information about the mapped environment (\eg objects (instances) and their classes) is retrieved from the robot's sensor data.
    \item[3. Prior knowledge integration.] The geometric and semantic information from sub-problems 1 and 2 is connected to existing prior knowledge, \eg via an ontology. Additional semantic information (\eg the room types) is then derived using reasoning.
\end{description}

This decomposition of the mapping process provides a modular and flexible architecture for a semantic mapping system since components can be replaced easily depending on current task requirements, but it has several critical shortcomings. One key shortcoming stems from the rigid flow of the steps, in which the semantic mapping process is divided into separate sub-problems that are handled consecutively due to dependencies on earlier tasks (the prior knowledge integration, for example, makes no sense without semantic information from the actual environment). The geometric mapping and semantic information steps therefore have no access to and do not benefit from the prior knowledge brought in at the last step, even though this prior knowledge could be used to correct mistakes made in those tasks, \eg classification errors.

Recent advances in deep learning have enabled the integration of symbolic data represented as, \eg knowledge graphs, into models utilizing sub-symbolic embeddings generated directly from sensor data (images, point clouds) \citep{Gouidis2019-dk}. In this field, 
\textbf{semantic scene graphs} and the integration of \textbf{(vision) language models} are of particular interest, as they enable models to use existing graph-based prior knowledge or to utilize the reasoning capabilities of large language models trained on internet-scale datasets. This survey aims to provide a comprehensive overview of recent developments in these fields, focusing on the online integration of prior knowledge into semantic data acquisition methods as well as semantic mapping as a whole. 

The remainder of the paper covers the classical approaches before discussing the new methods and the future directions of the field. Section~\ref{sec:nutshell} provides a more detailed overview of the classical semantic mapping steps and outlines their respective goals and requirements. In Sections~\ref{sec:scene_graphs} and \ref{sec:language}, we then review and analyze the recent developments and research trends in 3D semantic scene graphs and (vision) language models, including a discussion of open challenges for applications in semantic mapping. 

\section{Semantic mapping \textit{in a nutshell}}
\label{sec:nutshell}


As introduced above, semantic mapping as a whole can be roughly decomposed into geometric mapping, acquisition of semantic information from sensor data, and the integration of prior knowledge \citep{hanSemanticMappingMobile2021}. Each sub-problem is vital on its own (\eg geometric maps can be used immediately for tasks such as navigation even without any semantic information) and a useful semantic map requires all three, necessitating a sound understanding of each component. We therefore explain each sub-problem in depth in this section and outline popular solutions to each task.

\subsection{Geometric mapping}
\label{subsec:geometry_mapping}

A geometric representation of basic structures is a critical foundation for any useful map of an environment. In robotics, these geometric maps are typically built incrementally using \textbf{Simultaneous Localization and Mapping (SLAM)} approaches, which rely on the robot's built-in sensors to fuse noisy sensor data into a consistent geometric representation.

In earlier works, these maps were typically built using laser scanners and represented as occupancy grids in 2D \citep{grisetti2007improved}. Later advances then allowed data obtained from 3D laser scanners in a stop-scan-go fashion to be registered in a globally consistent 3D point cloud using the Iterative Closest Point (ICP) algorithm, as well as optimization based on neighboring scan poses (\textit{loop closure}) to eliminate accumulated pose errors \citep{nuchter20076d}. These large 3D point clouds can be reconstructed into compact 3D meshes \citep{wiemann2018surface} to be used for navigation \citep{putz20163d}, for building object-centric maps \citep{gunther2017model}, or for generating other representations such as 3D octrees \citep{hornung2013octomap}. However, due to the large size of the raw point clouds, this processing is done primarily offline, whereas a fully functional autonomous robot requires the ability to complete such mapping in an online and adaptive manner.

LOAM~\citep{Zhang2014loam} sparked the development of a variety of efficient feature-based methods for online odometry estimation and mapping~\citep[\eg][]{shan2018lego, Shan2020lio-sam} on LIDAR data. In contrast, KISS-ICP~\citep{Vizzo2023kiss-icp} recently developed a system based on traditional point-to-point ICP with adaptive thresholding based on voxel grids for efficient odometry estimation.
The growing availability of affordable RGB-D cameras, which combine the visual data from a regular RGB camera with depth information, has further led to the development of 3D representations that enable the creation of 3D maps in real-time. KinectFusion \citep{izadi2011kinectfusion, whelan2012kintinuous}, for example, popularized the Truncated Signed Distance Function (TSDF) as a fast intermediate representation during mapping from which a 3D mesh can be derived efficiently as a post-processing step. Furthermore, derived approaches like ElasticFusion \citep{whelan2015elasticfusion}, which use different representations (points or surfels), have also been proposed under the same general principle as KinectFusion. 

With the development of visual-inertial SLAM approaches such as ORB-SLAM3 \citep{Campos2021orbslam}, DSO \citep{Engel2018dso}, VINS-Mono \citep{Qin2018vinsmono}, and VINS-Fusion \citep{Qin2018vinsfusion}, it has become feasible to acquire the necessary pose information for map integration in real-time using a monocular or stereo camera and an inertial measurement unit. Recently, approaches using implicit neural-network-based representations based on Neural Radiance Fields (NeRF) \citep{mildenhallNeRFRepresentingScenes2020, rosinol2023nerf} or Gaussian Splatting \citep{kerbl20233d, matsuki2024gaussian, yan2024gs} have been proposed. However, these approaches require a heavy optimization step to generate the final representation, limiting their applicability in scenarios where the representation has to be frequently updated onboard a mobile robot.

\subsection{Semantic information from sensor data}
\label{subsec:acquision}

Once a robot has generated a geometric map of its environment via SLAM or other techniques, it needs to incorporate semantic information about that environment to move from a basic understanding of \textit{where} entities and structures are and what the general layout of its environment is, to an understanding of \textit{what} those entities and structures are and what allowances and conditions they present it with \citep{hanSemanticMappingMobile2021}. Broadly, there are two ways in which the robot can gain the information necessary to do this: bottom-up, by obtaining the information internally by looking for patterns and useful attributes in the sensor data it collects, or top-down, by referencing external knowledge \citep{Qiu2023-ou}.

To connect the geometrically mapped environment to a symbolic knowledge representation, a bridge between the sensor data and the explicit symbols has to be established. This way, the symbols can be \textit{grounded} in the map representation allowing a reasoning system to derive additional symbolic knowledge from the map. The first available data source is the 3D data from the geometric map itself. Using 3D models of known objects, \eg furniture \citep{gunther2017model}, it is possible to detect the poses of instances of these objects in the geometric map using algorithms such as RANSAC\@. From these instances, additional semantic knowledge about the spatial relations between the objects can be derived using spatial databases as done in the semantic environment mapping framework SEMAP \citep{deeken2015semap, deeken2018grounding}. In most works, the semantic information is derived directly from the sensor data provided by a robot's camera and 3D sensors, and advances in deep learning have enabled the development of powerful models for semantic and instance segmentation which allow this process to occur in real-time even on a self-contained mobile robot. 

Recent works often use Segment Anything~\citep{kirillov2023segment} and related models (e.g. MobileSAM~\citep{zhang2023fastersegmentanythinglightweight}, FastSAM~\citep{zhao2023fast}, SAM2~\citep{ravi2024sam2}) in order to obtain open-vocabulary or class-agnostic segmentation masks.
These masks are often projected into 3D space, enabling additional processing of detected objects, including clustering and detection of dynamic objects (such as humans or moving cars) that should not be included in the static map.
The resulting hybrid map representation can already be used as a semantic map for simple tasks, like avoiding forbidden areas (\eg sidewalks or bike lanes in autonomous driving). However, as no further semantic information is derived from additional prior knowledge, reasoning on such a map remains limited.

Today, a combination of geometric mapping and semantic segmentation is already provided by many semantic mapping frameworks such as Kimera~\citep{Rosinol2021-ko}, Voxblox++~\citep{Grinvald2019voxbloxplusplus}, PLVS~\citep{Freda2023plvs}, and the voxel-based semantic maps~\citep{Banegas-Luna2023voxelized} of the ViMantic robotic architecture~\cite{Fernandez-Chaves2021vimantic}. The recently released Khronos~\citep{Schmid-RSS24-Khronos} approach defined the \textit{Spatio-Temporal Metric-Semantic} SLAM (SMS) problem and provided a framework to solve it based on a factor graph, additionally considering dynamics and the evolution of scenes. 

A closely related problem to SMS, especially in the robotics field, is \textit{anchoring}~\citep{Coradeschi2003}. Anchoring seeks to establish direct connections between object instances detected in the sensor data by the aforementioned or other methods, and the symbols provided by a knowledge base. However, while the approaches outlined before are focused on the static parts of the environment, an anchoring system mainly handles \textit{dynamic} objects, \eg for re-discovery of already known object instances that have subsequently been moved ~\citep{Guenther2018ras}.

The amount of knowledge a robot can glean about the world around it simply by analyzing sensor data is surprisingly extensive. Without drawing on external knowledge, it can already generate basic information about the locations of distinct entities and their shapes or other properties, hierarchical information about the objects in that environment, and relative physical distances, which may reflect conceptual relations. The physical world also has a highly hierarchical structure, which can be discovered simply by observing it \citep{Feng2023-vh}. 

Analysis of sensor data taken inside a house, for example by feeding the semantic class labels through a graph autoencoder to develop class-dependent representations and an understanding of the topography of the captured scenes \citep{Zhang2021-zc}, will already yield information such as the fact that cups usually rest on tables and cushions on couches and not vice versa. More general rules, such as smaller objects generally appearing on top of larger ones, can be extracted even without identifying the objects in the environment, in the case that no external knowledge has been made available in any form (which is technically not the case when a trained object detector is used, as the detector may receive external knowledge during training in the form of provided labels). A robotic system designed to identify and leverage features in its environment based on sensor data can already significantly refine and constrain the predictions it makes about its environment, although this information is not sufficient for many tasks \citep{Qiu2023-ou,Feng2023-vh}. 

Deriving semantic information from the environment based on sensor data processing already enables a robot to enhance its reasoning and planning, and presents potential benefits for perception as well. However, key information about objects in the environment often cannot be gained from processing the sensor data alone, and in some cases, this basic processing can even lead to erroneous conclusions. For example, a wheelbarrow and a spade might be physically located at some distance from each other in a garage, but these are in fact highly related objects that both belong to the general topic of gardening. Such information must be obtained from external sources of knowledge to provide the robot with a complete semantic understanding of its environment, which is vital if the robot is to complete meaningful tasks alongside or in assistance of a human \citep{lang2014semantic}. How precisely to provide and incorporate this information into a robot's systems has long been a key focus of research, and continues to be a highly active topic today.  

\subsection{Prior knowledge integration}
\label{subsec:knowledge_integration}

For a robot to leverage knowledge in its actions in a specific environment, that knowledge must first be organized and made available to the robot in a suitable form. Of particular importance to a robot are its own capabilities and parameters, the characteristics of the entities in its environment, their properties and relationships with each other, and viable actions in a given state \citep{tenorth_knowrob_2013,cornejo-lupa_survey_2021}. Organizing this information is a non-trivial task, particularly given that both the sensor data for an object to be searched for in a knowledge base and the knowledge base itself are inevitably incomplete \citep{Qiu2023-ou}. Nonetheless, many solutions to this task have proven promising, with the most prominent methods being systems of predicates and rules \citep[][]{wang_semantic_2020,liu_generalizable_2020}, formalized in ontologies \citep[][]{cornejo-lupa_ontoslam_2021} and knowledge graphs \citep[][]{kalanat_symbolic_2022}.

Ontologies 
are carefully curated, code-form sets of concepts (classes) and their individual properties and relationships to each other. 
Despite being developed nearly three decades ago, the Resource Description Framework (RDF) for representing information and the description-logic-based Web Ontology Language (OWL) \citep{doing2024towards} remain the most prominent systems for describing ontologies even in modern applications, including robotic systems \citep{cornejo-lupa_survey_2021}. RDF and OWL standardized the basic structure of an ontology, greatly contributing to their popularity and success, although the newer field of knowledge graphs has shown equal promise. 

Where ontologies focus on a hierarchical structure to describe classes, knowledge graphs are mostly concerned with the relationships between them. Although several different formalizations of KGs exist,
there is an implicit consensus on their basic structure. Entities are typically represented as nodes in the graph while edges represent relationships between the entities \citep{Ji2022-xc}. Such graphs can be curated from multiple sources and allow for easy use by systems needing to reason on the contained knowledge. 



IBM’s Watson system deployed knowledge graphs and
showcased their suitability for integration into complete systems able to process incoming language, access relevant knowledge based on the incoming text, and provide appropriate answers in real time \citep{ferrucci_building_2010}. More recent 
methods take advantage of a variety of general knowledge sources produced since the late 2000s, including key collections like WordNet \citep{Miller1995wordnet}, Visual Genome \citep{Krishna2017-fh}, and ConceptNet \citep{Speer2017-os}. 

These large-scale graphs provide information on a vast range of topics for use in knowledge-integrated systems, perhaps most prominently in the field of visual data and image processing \citep[\eg][]{kalanat_symbolic_2022}. Recent works 
have increasingly partnered the knowledge graph with some form of graph neural network to take advantage of the inherent graph structure of these networks to effectively incorporate large knowledge graphs into visual classification pipelines, allowing information on known object classes to assist in the identification of unfamiliar objects \citep{wang_zero-shot_2018,lee_multi-label_2018}. However, the scale of many KGs poses a challenge for many fine-grained applications.

The vast information available in many commonsense knowledge graphs prevents their focused application to many robotics tasks, which require the robot to be supplied with just the tailored information it needs for the task at hand. Progress on this point has been made with more specialized ontologies, such as the OntoSLAM ontology designed specifically for SLAM applications \citep{cornejo-lupa_ontoslam_2021} and the KnowRob ontology, a more general ontology aimed at supporting a variety of robotics tasks \citep{tenorth_knowrob_2013,beetz_know_2018}. Such specialized ontologies already address the scale and applicability issue of large knowledge graphs, but work in this direction will continue to be an active and vital subfield for both ontologies and knowledge graphs. 

Ontologies require care when entries are updated or added \citep{noy2004ontology}, but they benefit from decades of interest and a clear consensus on definition and structure, as well as the curation of multiple general ontologies. Likewise, the propagation of information through the graph structure of knowledge graphs must be carefully managed to avoid information loss \citep{kampffmeyer_rethinking_2019}, but they are well-suited to modern neural network approaches. Both methods separately, and especially hybrid approaches combining them, have proven to be effective backbones for systems leveraging prior knowledge in the creation of semantic maps, and other robotics tasks. Particularly with the advent of large language models (LLMs), knowledge stored in knowledge graphs and ontologies can be accessed more efficiently \citep{Lewis2020-qd} and represented utilizing the implicitly stored common sense knowledge \citep{Strader2023-je,bosselut_comet_2019} (see Section~\ref{sec:language}). These approaches are and will likely remain key methods of knowledge representation in and beyond the field of robotic semantic mapping.


\section{3D semantic scene graphs}
\label{sec:scene_graphs}


The requirement for environment representations with richer semantics for robotic applications brought 3D semantic scene graphs into focus in the field. These graphs abstract the environment into a representation where nodes and the edges (connections) between them represent physical entities and their spatial or logical relations to each other \citep{Armeni2019-ip, Rosinol2020-wi, Hughes2022-fg, Kim2020-we, Wald2020-yj, Qiu2023-ou, Kabalar2023-mb}. 

3D scene graphs can broadly be differentiated into two categories based on their structure \citep{Bae2023-ai}. Those taking the form of a layered graph in which different layers represent different levels of abstraction (\eg mesh-objects-rooms-building) are deemed \textit{hierarchical}, while those taking a simpler form in which the graph represents only physical entities and their relations without inferring any hierarchy to more abstract concepts are deemed \textit{flat}. 
Scene graphs as an abstract representation are neither novel nor exclusive to robotics. As a familiar and vital tool in computer graphics, they are common today in many image processing tasks like visual question answering \citep{Teney2017-fm} and image captioning \citep{Yao2018-cb}. Established datasets, most prominently the Visual Genome dataset \citep{Krishna2017-fh}, have supported the development of numerous learning-based and other approaches for scene graph construction and processing. Semantic mapping for robotics, however, requires moving beyond those.

In contrast to the earlier, more basic scene graphs, 3D scene graphs are representations of whole scenes and have to be generated from interrelated sensor data rather than from a single image. Generating robust 3D scene graphs from sensor data is non-trivial, with many different methods spawned in recent years, broadly falling into two categories of approaches:  

\begin{itemize}
  \item \textit{Construction}: the nodes are detected in the sensor data, and the scene graph is subsequently built deterministically, meaning edges are inferred based on geometric properties like position and size. These methods commonly result in hierarchical scene graphs, since higher abstraction layers can be incrementally built (see Fig.~\ref{fig:rosinol-sg}). 
  \item \textit{Prediction}: the graph's topology is based on a heuristic, such as the proximity of objects to each other, and the classes of edges and nodes are predicted by \eg a graph neural network. Scene graph prediction usually results in flat scene graphs as prediction applies the sensor readings directly, and can thus be learned end-to-end (see Fig.~\ref{fig:flat-sg}).
\end{itemize}




It is generally argued that in robotics, hierarchical scene graphs have more potential since modeling an environment with levels of higher abstraction benefits planning \citep{Agia2022-mi,Rosinol2020-wi}. However, scene graph construction levels usually only infer spatial or geometric relations (\eg on, next to, bigger/smaller than), while more nuanced semantics (\eg connected to, standing on, hanging on) are harder to generate. These kinds of relations require a deeper understanding of the involved objects and the semantics behind their relations. 

While inferring these relations deterministically is cumbersome, learning them from labeled datasets like the 3DSSG dataset \citep{Wald2020-yj} is more viable. As is typically the case in deep learning, the ability of models for scene graph prediction to extract these relationships is highly dependent on dataset quality and availability, though modern models no longer need to rely exclusively on these sources. In addition to datasets like Visual Genome and 3DSSG, prior knowledge from sources like commonsense knowledge graphs, \eg WordNet \citep{Miller1995wordnet} or ConceptNet \citep{Speer2017-os}, can be utilized for inferring complex relationships. The same can be said for language models as knowledge sources, since the relations in question are also represented in our everyday language, as covered in depth in Section~\ref{sec:language}. Their potential has been demonstrated \citep{Gu2019-ul} and is currently emerging in 3D scene graph prediction as well. 



Generating scene graphs in 3D is challenging, since ambiguity in the data and incompleteness of the graph itself need to be addressed. However, 3D scene graphs can efficiently connect the subsymbolic and symbolic representations of an environment, providing a framework that generally improves high-level task planning, reasoning on semantic maps, and 
even path planning compared to geometric representations. 
In the following sections, we will discuss 3D scene graph generation methods further and include the consideration of prior knowledge in the generation process. We also introduce multiple application scenarios for 3D scene graphs from task planning, navigation, robot collaboration, and change detection.

\subsection{3D scene graph generation}
\label{subsec:scene_graph_generation}

3D scene graph generation includes methods that take 3D data, such as depth images or point clouds, and output an abstract graphical representation in the form of a scene graph. As discussed in the introduction of Section~\ref{sec:scene_graphs}, 3D scene graph generation models can be roughly divided into \textit{construction} and \textit{prediction} methods. For a compact overview of the presented methods and applications, see Tables~\ref{tab:sg_h} and \ref{tab:sg_flat}.

The first notion of a 3D scene graph generated from 3D data arose at the end of the last decade, when \citet{Armeni2019-ip} introduced a hierarchical scene graph consisting of 4 layers: a building level (as a root node), a room level, an object level, and a level for the camera poses from which the data was captured. The graph itself is constructed by projecting Mask R-CNN \citep{He2017-qx} results from 2D images onto a 3D mesh, with two heuristics used to resolve ambiguities in the segmentation by aggregating multiple views of objects in panoramic images. The derived relationships from this method were still mainly geometric, namely left/right, in front of/behind, bigger/smaller, occlusion relationships, and relationships to other layers, but this framework provided a key foundation for later advances. 

\begin{figure}[t]
    \centering
    \includegraphics[width=\linewidth]{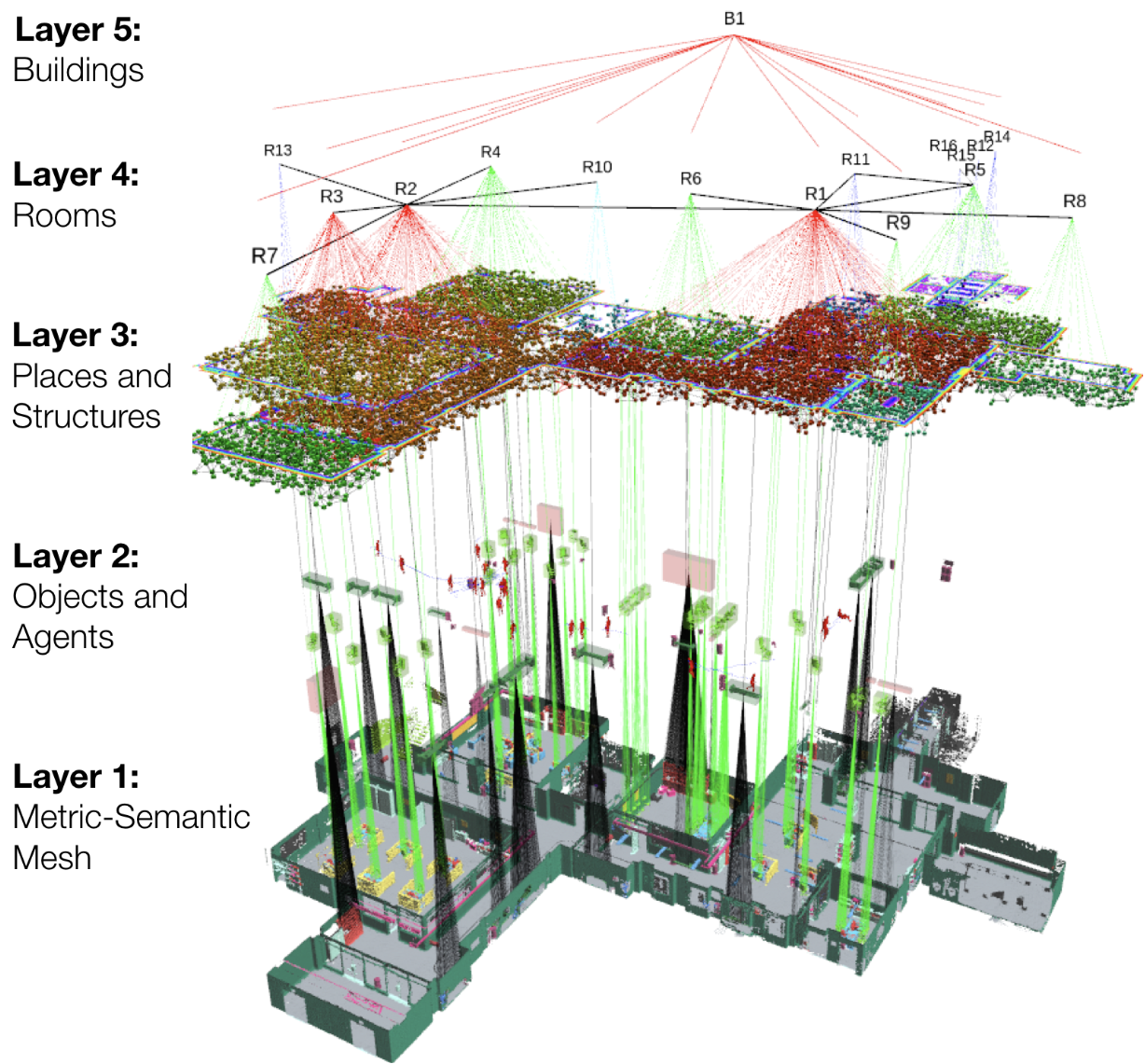}
    \caption{Hierarchical scene graph from \cite{Rosinol2021-ko}}
    \label{fig:rosinol-sg}
\end{figure}

Improving on the idea of hierarchical scene graphs, \citet{Rosinol2020-wi} presented a method for automatic 3D scene graph construction from visual-inertial data. The described 3D dynamic scene graph (DSG) comprises 5 layers (see Figure \ref{fig:rosinol-sg} for a visual representation):

\begin{enumerate}
    \item The metric semantic layer with labeled vertices and edges forming faces
    \item Objects (static non-structures) and dynamic agents (\eg humans and robots) extracted from the mesh
    \item Places (free, navigable spaces) and structures (walls, floors, ceilings, pillars)
    \item Rooms, corridors, and halls
    \item Building (as the root node)
\end{enumerate}

In a DSG, nodes can be connected within the same layer or across different layers, with edges within a layer representing spatial relations like contact, adjacency, or traversability, and edges between layers forming the explicit hierarchy and allocating lower-layer entities to higher-layer ones (objects to places, places to rooms, etc.) DSGs are constructed using the Spatial Perception Engine (SPIN), based on the metric-semantic mesh and the ESDF generated by Kimera~\citep{Rosinol2021-ko}. Objects and structures are thus estimated from the labeled mesh, while humans (representative for agents) are detected using a Graph CNN\@. Places are sampled from the free space in the ESDF and rooms are then estimated from these sampled places.

Building on DSGs and SPIN, Hydra \citep{Hughes2022-fg} adds real-time capability and loop-closure to run on an operating robot. This more advanced framework includes improved algorithms for mesh and place construction, as well as an improved algorithm for room detection. The loop closure detection is based on hierarchical descriptors comparing the scene appearance, present objects, and places. On detected loop closure, the DSG is then optimized by correcting the drift and merging overlapping nodes.

In contrast to scene graph construction, scene graph prediction requires high-quality labeled datasets, ideally containing 3D data and the associated scene graph, to train a neural network. The Visual Genome dataset \citep{Krishna2017-fh} has been the base for many image-language-focused learning tasks like visual question answering and scene graph prediction, but is not optimal for training scene graph prediction networks as it only contains 2D images that are not necessarily linked to coherent scenes. A popular dataset that includes the 3D data needed for end-to-end learning for these networks is the 3D Semantic Scene Graph (3DSSG) dataset \citep{Wald2020-yj}, an extension of the 3RScan dataset \citep{Wald2019-ki} which consists of manually labeled scene graphs of indoor environments with 534 nodes and 40 edge classes based on RGB-D images, with a subset of 160 nodes and 27 edge classes used for training and validation. 

The initial graph topology for a 3D scene graph is usually derived from the geometric structure of the point cloud, oftentimes using instance segmentation \citep{Wald2020-yj, Wu_undated-qv, Li2022-ni} or clustering approaches \citep{Qi2024-zp}. The resulting point clusters then become nodes in the graph. Edges can be added either in a fully connected paradigm where every node is connected to every other node, or with heuristics targeting only useful or sensible edges, \eg creating an edge between two nodes whose point segments are within a certain distance of each other. 

For the actual prediction of node and edge classes, some form of graph neural network (GNN) architecture is typically used because the underlying message-passing mechanism in these networks allows for the aggregation of information from neighbor nodes in the graph. \Citet{Wald2020-yj} use a graph agnostic graph convolutional network with two layers to predict node classes from object points and edge classes from object pair points, while \Citet{Qi2024-zp} use Gated Recurrent Units \citep{Cho2014-sh} for this purpose. In both approaches, PointNet \citep{Qi2017-wl} is used to extract visual features from segmented object points for object classification and object pair points for edge classification. However, while such non-incremental scene graph generation methods can be utilized in semantic mapping, they are not optimal for dynamic environments, in which approaches that build the graph incrementally based on new sensor data are needed to allow for real-time updates and more accurate representations of the current scene. 

\begin{figure*}
    \centering
    \includegraphics[width=1\linewidth]{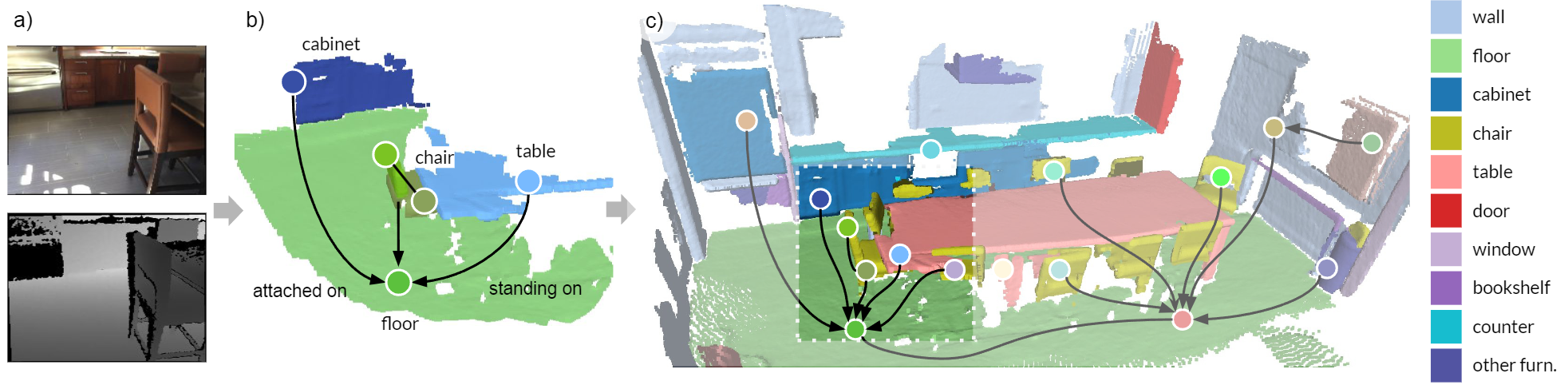}
    \caption{Flat incremental scene graph from \cite{Wu_undated-qv}}
    \label{fig:flat-sg}
\end{figure*}

One such approach is SceneGraphFusion \citep{Wu_undated-qv}, a pipeline specifically designed for incremental scene graph prediction on RGB-D data. Similar to the approach by \citet{Wald2020-yj}, the pipeline takes segmented point clouds as input, although it works on objects only visible in the current frame. A neighborhood graph is constructed by connecting objects or nodes within a fixed radius. In addition to PointNet features, simple geometric properties 
are derived as well. The pipeline employs a GNN, enhanced by a novel feature-wise attention mechanism (FAT) to manage incomplete 3D data and dynamic edges. Predictions focus on objects within the sensor frame, updating newly appearing objects and excluding old ones before inference, with the resulting nodes and edges integrated into a global scene graph through a running average approach (see Figure \ref{fig:flat-sg}).  

Similarly, \citet{Li2022-ni} developed a framework for embodied scene graph generation where a local scene graph is predicted on RGB-D images and merged with a global scene graph, endorsing free exploration while avoiding reliance on fixed paths. More recently, an approach was proposed using only RGB image sequences as input \citep{Wu2023-dm}, with the incremental framework using ORB-SLAM3 \citep{Campos2021orbslam} to estimate sparse 3D point clouds from RGB image sequences.

\subsection{3D scene graph prediction with prior knowledge}
\label{subsec:scene_graph_prior_knowledge}

As in visual reasoning or visual question answering, utilizing prior knowledge also has substantial potential to boost prediction performance with 3D scene graphs \citep{Gouidis2019-dk}. The applicability of such methods to 3D data is obvious and inspires current research in scene graph prediction. Knowledge graphs and ontologies are great representations for this purpose, since the graph structure integrates well with the commonly used graph neural network architectures and the scene graph representation in general. 

\citet{Feng2023-vh} utilize a hierarchical knowledge graph based on ConceptNet that includes common objects found in 3D point cloud scenes. Based on this hierarchy, a visual graph is constructed from the point cloud data. The visual graph and the knowledge graph are then passed into two different graph neural networks which embed the respective features with message passing before object detection and scene graph prediction are performed on these embeddings by an MLP and a GCN.

\Citet{Qiu2023-ou} integrate a knowledge graph based on WordNet, ConceptNet, and Visual Genome which models pairwise relationships. Using a Knowledge-Scene Graph Network (KSGN) based on the Graph Bridging Network \citep{Zareian2020-kc}, the knowledge graph is included in the message passing process in the scene graph, improving scene graph prediction on the 3DSSG160 dataset.

In addition to knowledge graphs, large language models can now be used as a source of external knowledge. \Citet{Lv2024-pn} present SGFormer, a transformer-based graph neural network that incorporates text embeddings from descriptions about known objects, which are generated using an LLM and integrated using cross-attention. The resulting architecture shows great improvements on the 3DSSG dataset, showcasing the usefulness of prior knowledge from LLMs for scene graph prediction. 

While most current 3D scene graph generation methods are demonstrated on indoor datasets, \citet{Strader2023-je} show an approach generating hierarchical 3D scene graphs for indoor and outdoor environments. In their graph hierarchy, they differentiate between high-level concepts like rooms, roads, and beaches, and low-level concepts like objects and places. Using an LLM, these concepts are then connected in a bipartite spatial ontology, which is incorporated using a neurosymbolic Logic Tensor Network \citep{Badreddine2020-yd}. During training, the satisfaction of axioms comparing the prediction to the ground truth and the spatial ontology is used as the loss. \Citet{Strader2023-je} demonstrate their approach on the Matterport3D dataset \citep{Chang2017-bk} and two outdoor datasets, showing an improvement over purely data-driven prediction when training on fewer samples.

Besides classical 3D scene graph generation, \citet{Giuliari2023-sw} introduce Directed Spatial Commonsense Graphs (D-SCG), heterogeneous scene graphs with relative spatial edges and commonsense nodes and edges obtained from ConceptNet. The commonsense nodes in these graphs are linked to spatial object nodes based on the ConceptNet predicates AtLocation and UsedFor, with the object nodes fully connected by directed proximity nodes containing the relative position vector between two objects. The D-SCG is used to predict the position of unseen objects in incomplete ScanNet \citep{Chang2017-bk} scenes.

Although not technically working with prior knowledge, another notable work in this area is \citet{Zhang2021-zc}, which proposes a method based on knowledge-based meta embeddings, i.e. embeddings of the one-hot encoded class vectors. The meta embeddings are integrated in a two-iteration prediction where the first iteration predicts node and edge classes purely on the point cloud data and the second integrates meta embeddings as features for the respective node and edge classes based on the classification from the first iteration. Similarly, \citet{Han2023-ve} propose unbiased meta embeddings, which are weighted by their appearance frequency in the 3DSSG dataset.  

\subsection{Scene graph applications}
\label{subsec:scene_graph_applications}

As a unifying structure between low-level geometric and high-level symbolic representations, 3D scene graphs appear in multiple application domains in robotics. \citet{Rosinol2020-wi} implement the places layer in DSGs for navigation, a topological subgraph representing free, navigable spaces in the map (see section  \ref{subsec:scene_graph_generation}) and refer to higher layers for high-level planning. These higher layers can be used to learn navigation policies \citep{Ravichandran2022-vd} for object-centric navigation. \citet{Sunderhauf2019-dh} also learns object-centric navigation policies, differentiating between landmarks (static objects) present in the graph and targets (dynamic objects) not in the current graph. Using a graph convolutional network, a probability distribution of the targets based on the landmarks is learned. 

\Citet{Li2022-ni} use imitation learning and reinforcement learning to predict a robot's next navigation action to explore an unknown environment for scene graph generation. The learned policy network is based on LSTMs and includes the last action, the current sensor frame, and a local and global scene graph to predict the next action from a discrete action space. \citet{Lingelbach2023-ak} learn navigation policies for hierarchical relational object navigation, enabling navigation policies for large hierarchical, multi-room environments. The proposed approach uses the Heterogeneous Graph Transformer \citep{Hu2020-zk} and a task-driven attention mask to embed a hierarchical scene graph with rooms and objects for policy prediction. In addition to object-centric navigation, 3D scene graphs can also be used in task planning, such as in \Citet{Amiri2022-de}, where a POMDP is used with scene graph biased belief in an object search scenario.

3D scene graphs have also been applied to change detection. Variable Scene Graphs \citep{Looper2023-kb}, for example, add the likelihood for an object to change in the future, either by state (\eg open/closed), 3D position, or instance (position in the graph topology). Scene graphs have also been used in visual localization based on RGB images by making image retrieval and pose estimation robust to dynamic objects  \citep{Kabalar2023-mb}. For this, scene graphs are estimated for a given image sequence and matched to reference scene graphs from an image database. Comparison and connection between frames and pose estimation is then carried out using a set of pre-defined classes which are unlikely to change.

While an intuitively clear structure, scene graphs can reach sizes at which they start to pose challenges to task planning in terms of run time and completion. For this, the task-planning framework Taskography \citep{Agia2022-mi} has been proposed to prune and sparsify large 3D scene graphs to decrease planning times and increase success rates. An algorithm to compress local scene graphs for communication (D-Lite) has also been proposed for multi-robot collaboration \citep{Chang2023-oc}. 

\begin{table*}[tb]
\small

\renewcommand\theadalign{l}
\renewcommand\cellalign{cc}

\begin{tabularx}{\textwidth}{p{2cm} p{2cm} X p{2.5cm} p{3.0cm}}
\toprule

\thead{Name} & \thead{Reference} & \thead{Generation Method} & \thead{Dataset} & \thead{Application} \\

\midrule


3D Scene Graph & \Citet{Armeni2019-ip} & \makecell*[{{p{4.5cm}}}]{Predicted on-demand by multiple models} & \makecell*[{{p{2.5cm}}}]{Gibson\\ Environment~\citep{Xia2018-qq}} & \makecell*[{{p{3cm}}}]{Mapping} \\
  \midrule
3D Dynamic Scene Graph & \Citet{Rosinol2020-wi} & \makecell*[{{p{4.5cm}}}]{Explicit offline construction; known dynamic classes (\eg humans) filtered} & \makecell*[{{p{2.5cm}}}]{uHumans~\citep{RosinolUnknown-hi}} & \makecell*[{{p{3cm}}}]{Mapping} \\
  \midrule
Hydra & \Citet{Hughes2022-fg} &  \makecell*[{{p{4.5cm}}}]{Explicit online construction} & \makecell*[{{p{2.5cm}}}]{uHumans2~\citep{Rosinol2021-ko},\\ SidPac~\citep{Hughes2022-fg}} & \makecell*[{{p{3cm}}}]{Mapping} \\ 
  \midrule
Taskography & \Citet{Agia2022-mi} & \makecell*[{{p{4.5cm}}}]{3D Scene Graph~\citep{Armeni2019-ip}} & \makecell*[{{p{2.5cm}}}]{Gibson\\ Environment} & \makecell*[{{p{3cm}}}]{Task planning} \\ 
  \midrule
D-Lite & \Citet{Chang2023-oc} & \makecell*[{{p{4.5cm}}}]{Method from~\citep{Rosinol2021-ko}} & \makecell*[{{p{2.5cm}}}]{uHumans2} & \makecell*[{{p{3cm}}}]{Scene graph compression} \\
  \midrule
  - & \citet{Feng2023-vh} & \makecell*[{{p{4.5cm}}}]{Offline prediction with hierarchical symbolic knowledge} & \makecell*[{{p{2.5cm}}}]{3DSSG} & \makecell*[{{p{3cm}}}]{Predicate prediction}\\
  \midrule
 - & \citet{Strader2023-je} & \makecell*[{{p{4.5cm}}}]{Hydra~\citep{Hughes2022-fg}} & \makecell*[{{p{2.5cm}}}]{MP3D~\citep{Chang2017-bk}, custom outdoor dataset} & \makecell*[{{p{3cm}}}]{Prediction of place labels}\\
  \midrule
  
HREM & \Citet{Ravichandran2022-vd} & \makecell*[{{p{4.5cm}}}]{Method from \citep{Rosinol2020-wi}} & \makecell*[{{p{2.5cm}}}]{Office simulation} & \makecell*[{{p{3cm}}}]{Object-centric navigation} \\
  \midrule

HRON & \Citet{Lingelbach2023-ak} & \makecell*[{{p{4.5cm}}}]{Constructed from simulation} & \makecell*[{{p{2.5cm}}}]{iGibson 2.0\\ environment~\citep{Li2022-ps}} & \makecell*[{{p{3cm}}}]{Hierarchical relational object navigation} \\

\bottomrule
\end{tabularx}
\caption{Overview and comparison of references by the method used for hierarchical scene graph generation, the dataset used for benchmarking, and the application presented in the paper (if not only the generation method)}\label{tab:sg_h}

\end{table*}

\begin{table*}[tbp]
\small

\renewcommand\theadalign{l}
\renewcommand\cellalign{cc}
\begin{tabularx}{\textwidth}{p{2cm} p{2cm} X p{2.5cm} p{3.0cm}}
\toprule

\thead{Name} & \thead{Reference} & \thead{Generation Method} & \thead{Dataset} & \thead{Application} \\

\midrule

\makecell*[{{p{2cm}}}]{3-D\\ scene graph} & \Citet{Kim2020-we} &  \makecell*[{{p{4.5cm}}}]{Image-level prediction using\\ Factorizable Net} & \makecell*[{{p{2.5cm}}}]{ScanNet~\citep{Dai2017-jz}} & Task planning, VQA\\

\midrule 
  
3DSSG & \Citet{Wald2020-yj} &  \makecell*[{{p{4.5cm}}}]{Offline Prediction with GNN} & Modified 3RScan~\citep{Wald2019-ki} & Prediction\\

\midrule 
  
\makecell*[{{p{2cm}}}]{SceneGraph-\\ Fusion} & \Citet{Wu_undated-qv} &  \makecell*[{{p{4.5cm}}}]{Online incremental prediction from single frames with GNN} & 3DSSG~\citep{Wald2020-yj}, ScanNet \citep{Dai2017-jz} & Incremental Prediction\\

\midrule 

MonoSSG & \Citet{Wu2023-dm} &  \makecell*[{{p{4.5cm}}}]{Online incremental prediction from RGB sequences with GNN} & 3DSSG & Incremental Prediction\\

\midrule 

KSGN & \Citet{Qiu2023-ou} & \makecell*[{{p{4.5cm}}}]{Offline prediction of scene graphs with bridging to knowledge graphs} & 3DSSG & Prediction\\

\midrule 

SGFormer & \Citet{Lv2024-pn} & \makecell*[{{p{4.5cm}}}]{Offline prediction with transformer model and text embeddings} & 3DSSG & Prediction \\

\midrule 

DGGN & \Citet{Qi2024-zp} &  \makecell*[{{p{4.5cm}}}]{Offline prediction on point clusters} & \makecell*[{{p{2.5cm}}}]{s3DIS~\citep{Armeni2016-ud}, \\ 3DSSG, \\Paris-Lille-3D~\citep{Roynard2018-mj}} & Prediction \\

\midrule 

LSSG+GSSG & \Citet{Li2022-ni} &  \makecell*[{{p{4.5cm}}}]{Online incremental prediction from single frames with GNN} & AI2Thor~\citep{Kolve2017-nz} & Navigation  \\

\midrule 

\makecell*[{{p{2cm}}}]{Object-based\\ SLAM} & \Citet{Sunderhauf2019-dh} &  \makecell*[{{p{4.5cm}}}]{Construction of random pose- and landmark nodes} & \makecell*[{{p{2.5cm}}}]{Custom indoor\\ scenario} & \makecell*[{{p{2.5cm}}}]{Object-centric navigation} \\

\midrule 

SARP & \Citet{Amiri2022-de}  &  \makecell*[{{p{4.5cm}}}]{Predictions on 360-degree RGB images} & \makecell*[{{p{2.5cm}}}]{Custom\\ environment,\\ AI2Thor} & Object search\\

\midrule  

3D VSG & \Citet{Looper2023-kb} &  \makecell*[{{p{4.5cm}}}]{GNN on ground truth scene graph} & 3DSSG, 3RScan & Change detection \\

\midrule 

- & \Citet{Kabalar2023-mb} &  \makecell*[{{p{4.5cm}}}]{Online construction from single images} & RIO10\citep{Wald2020-nw} & Visual localization \\

\midrule 

D-SCG & \citet{Giuliari2023-sw} & \makecell*[{{p{4.5cm}}}]{Offline construction from ground truth segmentations} & \makecell*[{{p{2.5cm}}}]{ScanNet, ConceptNet~\cite{Speer2017-os}} & \makecell*[{{p{3cm}}}]{Object localization in partial scenes} \\

\midrule 

KiSGPM & \citet{Zhang2021-zc}  & \makecell*[{{p{4.5cm}}}]{Offline prediction with knowledge-based meta-embeddings} & 3DSSG & Prediction \\
  
\midrule

MP-DGCNN + ENA-GNN & \citet{Han2023-ve} & \makecell*[{{p{4.5cm}}}]{Offline prediction with unbiased knowledge-based meta-embeddings} & 3DSSG & Prediction \\ 

\bottomrule
\end{tabularx}
\caption{Overview and comparison of references by the method used for flat scene graph generation, the dataset used for benchmarking, and the application presented in the paper (if not only the generation method) }\label{tab:sg_flat}

\end{table*}

\subsection{Challenges}

3D scene graphs are a promising representation for complex environments, but generating them still entails certain challenges. Creating and learning from a labeled scene graph dataset, for example, is difficult as scene graphs are by nature incomplete. A neural network may therefore predict relationships that are technically correct, but which are not present in the ground truth, hindering the use of common loss functions like cross-entropy on predicated edge labels. 
 
The issue of incompleteness also surfaces in the metrics used to measure prediction performance. As of now, the \textit{recall@k} \citep{Lu2016-mq} is commonly used, which tracks the fraction of correct predictions within all predictions. In recent studies, the \textit{mean recall@k} \citep{Chen2019-bh} has been gaining popularity as a method that can handle the imbalanced nature of the long-tailed predicate distributions in scene graph datasets. Both \textit{recall@k} and \textit{mean recall@k} are insensitive to false positives, i.e., both would not be influenced by predicted relationships that are not present in the ground truth. 

Models measured with precision or accuracy tend to show comparatively low scores because most datasets lack negative annotations and therefore only represent relationships that are correct and not those that are known to be wrong. Future datasets may add such negative annotations (such as a glass being annotated not only as standing on a table but also as \textit{not} standing under it), but for now, the issue of negative sampling remains a complex one in 3D scene graph prediction and other applications and is likely to be a key area of future research in the field. 

More complex applications for 3D scene graphs remain to be developed. At present, existing applications for 3D scene graphs such as \cite{Ravichandran2022-vd} and \cite{Sunderhauf2019-dh} are still limited in their action space and the semantic richness of the scene graphs used, with current work usually focusing on scene graph construction \textit{or} prediction and not the combination of both. Further work is needed to compare the performance of construction and prediction approaches and to explore their incorporation and combination into unified frameworks taking advantage of the benefits of both methods. 

\section{Language models}
\label{sec:language}

Traditionally, information gathering (Section~\ref{subsec:acquision}) and reasoning (Section~\ref{subsec:knowledge_integration}) rely on some form of closed set \eg a fixed set of classes obtained from a segmentation model. Low-shot methods, including few- and zero-shot methods, attempt to enable a model to distinguish sparsely seen or even unseen categories during inference \citep{huWhatCanKnowledge2022, renVisualSemanticSegmentation2023}. However, these models typically lack a method to assign a meaningful name to these novel categories, instead utilizing a set of \textit{semantic features}, such as the detected objects' color, shape, or composition, to categorize unseen classes.

\begin{figure*}[htb]
\centering
\includegraphics[width=.9\textwidth]{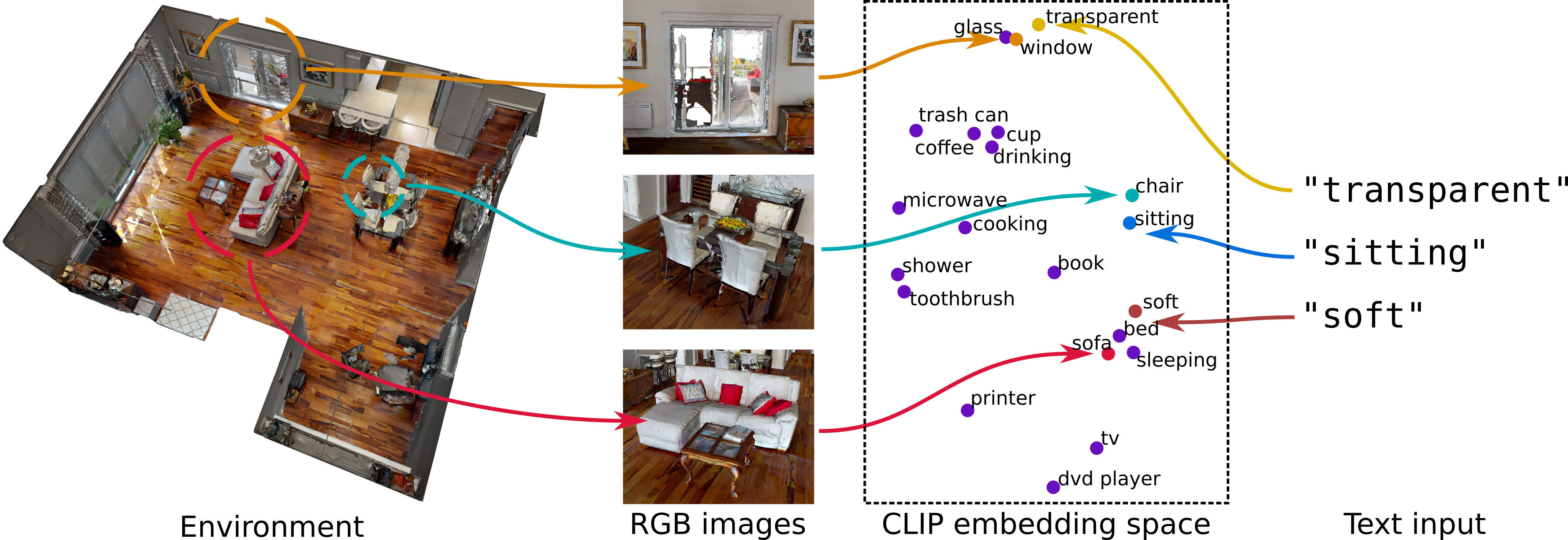}
\caption{Association of textual input with visual concepts and properties in a single embedding space using the CLIP model.}\label{fig:embedding_space}
\end{figure*}

By combining textual and visual data, \textit{Vision-Language Foundation Models} (VLFM) such as the pioneering \textit{Contrastive Language-Image Pre-Training} Model (CLIP) \citep{radfordLearningTransferableVisual2021} learn rich multi-modal features combining natural language with visual concepts in a common feature space (see Fig. \ref{fig:embedding_space}). Recent VLFM-based models excel in low-shot open-vocabulary object detection \citep{liu2023grounding, ren2024grounding} and segmentation tasks \citep{zhou2023zegclip}, with their performance matching or even exceeding that of traditional models. 

Recently, \textit{Large Language Models} (LLMs) have also emerged as a promising tool in these applications \citep{zeng2023large}. Training on vast text corpora consisting of billions of examples has enabled these models to process natural language inputs and generate high-quality natural language texts, sparking many innovations in natural language processing and other fields. By leveraging these models and integrating them with \eg VLFMs for processing visual input, the new family of \textit{Large Multimodal Models} (LMMs) now allows visual input from images or videos to be processed directly \citep{gao2024sphinx, team2023gemini, li2022blip, yang2023dawn, li2024llava}. 

Both VLFMs and LMMs allow for significant enhancements of both traditional and scene-graph-based semantic mapping approaches (see Section~\ref{sec:scene_graphs}) by combining explicit geometric and semantic scene representation with powerful language processing capabilities for both object representation and reasoning.

In the following section, we analyze the recent advances in VLFMs and LMMs and outline their impact on map representation and semantic mapping applications.

\subsection{Vision-language foundation models}
\label{subsec:vlm}

The most popular Vision-Language Foundation Model is presently CLIP \citep{radfordLearningTransferableVisual2021}, a joint text and image embedding model trained in a self-supervised paradigm using \textit{contrastive pre-training} on $\approx400$ Million text-image pairs, first introduced in 2021. CLIP maps images and textual input into an embedding space which maintains images and their matching captions proximate to each other, thus reducing inference to a simple neighbor search in the embedding space based on an input image and different textual descriptions.

The initial CLIP model already showed remarkable performance in zero-shot image classification tasks, even matching supervised models trained on benchmark datasets. This, and the model's comparative simplicity and straightforward integration into various downstream tasks, led to the growing popularity of VLFMs for many computer vision tasks. However, CLIP \citep{radfordLearningTransferableVisual2021} and similar baseline models like ALIGN \citep{jia2021scaling}, BLIP \citep{li2022blip}, and BLIPv2 \citep{li2023blip} only produce feature vectors for the \textit{entire} image. This prevents their direct applicability in the semantic and instance segmentation tasks required for semantic mapping, as these require fine-grained features to localize detected objects and other instances \textit{within} an image.

Pixel- and region-based approaches address this shortcoming \citep{zhou2022extract, ghiasi2022scaling, ding2022open, zou2023generalized, lilanguage, xu2023side, luddecke2022image} by implementing 2D open-vocabulary semantic segmentation for each pixel or larger regions of pixels. Using baseline VLFMs and recent instance-agnostic segmentation models \citep{kirillov2023segment} as foundations, they learn to generate individual feature vectors for each pixel or region in the input image, albeit at the cost of significantly higher run times in comparison to image-based VLFMs and traditional models \citep{yamazaki2024open}.



In contrast to traditional segmentation models, the models covered here do not require the segmentation classes to be provided at training time. Instead, the requested classes are provided to the model on demand for each image, allowing for queryable scene representations for arbitrary objects in semantic mapping tasks. However, to support 3D mapping tasks, the 2D segmentation capabilities of these models must first be lifted into the 3D domain, and, as the segmentation classes are not fixed during training, this cannot be achieved by simply projecting the resulting segmentation masks onto the 3D geometry as in established approaches \citep{Rosinol2021-ko}. Instead, this must be done by storing the features within the map itself. We discuss a multitude of recent approaches in this direction (summarized in Tab. \ref{tab:vlfm}) in the following sections.

\begin{table*}[tbp]
\small
\renewcommand\theadalign{l}
\renewcommand\cellalign{cc}
\begin{tabularx}{\textwidth}{p{2cm} p{2cm} p{2cm} X p{2cm} p{2cm}}
\toprule

\thead{Name} & \thead{Reference} & \thead{Map Type} & \thead{VLFM Integration} & \thead{Inference Type} & \thead{Application} \\

\midrule
LERF & \makecell*[{{p{2cm}}}]{\Citet{kerrLERFLanguageEmbedded2023}} & NeRF~\citep{mildenhallNeRFRepresentingScenes2020} & \makecell*[{{p{3.5cm}}}]{Averaging of CLIP~\citep{radfordLearningTransferableVisual2021} features from image pyramid} & \makecell*[{{p{2cm}}}]{View Poses \& Saliency maps} & Visualization \\

\midrule

CLIP-Fields & \makecell*[{{p{2cm}}}]{\Citet{shafiullah2022clip}} & NeRF & \makecell*[{{p{3.5cm}}}]{Detic~\citep{zhou2022detecting} segmentation \& CLIP features for each segment} & \makecell*[{{p{2cm}}}]{View poses} & Visualization \\

\midrule

OpenScene &  \makecell*[{{p{2cm}}}]{Peng and \\ Genova~\citep{pengOpenScene3DScene2023a}} & Point cloud & \makecell*[{{p{3.5cm}}}]{Distillation of OpenSeg~\citep{ghiasi2022scaling} to 3D point clouds} & \makecell*[{{p{2cm}}}]{Saliency maps \& Geometry} & Mapping \\

\midrule

ConceptFusion &  \makecell*[{{p{2cm}}}]{\Citet{jatavallabhula2023conceptfusion}} & Point cloud & \makecell*[{{p{3.5cm}}}]{Pixel-features from segment crops~\citep{kirillov2023segment, cheng2022masked} projected to 3D points} & \makecell*[{{p{2cm}}}]{Geometry \\segments} & Mapping \\

\midrule

OK-robot &  \makecell*[{{p{2cm}}}]{\Citet{liu2024ok}} & \makecell*[{{p{2cm}}}]{Voxelized\\ point cloud} & \makecell*[{{p{3.5cm}}}]{Similar to \citep{jatavallabhula2023conceptfusion}} & \makecell*[{{p{2cm}}}]{Top voxel} & \makecell*[{{p{2cm}}}]{Language-based\\ navigation \&\\ manipulation} \\

\midrule

Open-Fusion & \makecell*[{{p{2cm}}}]{\Citet{yamazaki2024open}} & TSDF mesh & \makecell*[{{p{3.5cm}}}]{Confidence maps of SEEM~\citep{zou2024segment} projected to 3D points} & \makecell*[{{p{2cm}}}]{Representative points} & Mapping \\

\midrule

HOV-SG & \makecell*[{{p{2cm}}}]{Werby\\ et al.~\citep{werby2024hierarchical}} & Hydra~\citep{Hughes2022-fg} & \makecell*[{{p{3.5cm}}}]{Similar to \citep{jatavallabhula2023conceptfusion} \& feature aggregation in object and parent nodes} & \makecell*[{{p{2cm}}}]{Object nodes} & \makecell*[{{p{2cm}}}]{Language-based\\ navigation} \\

\midrule

ConceptGraphs & \makecell*[{{p{2cm}}}]{\Citet{guConceptGraphsOpenVocabulary3D2023}} & \makecell*[{{p{2cm}}}]{Similar to\\ Khronos~\citep{Schmid-RSS24-Khronos}} & \makecell*[{{p{3.5cm}}}]{Similar to \citep{jatavallabhula2023conceptfusion} \& scene graph structure predicted by LLM} & \makecell*[{{p{2cm}}}]{Object nodes} & Mapping \\

\midrule

CLIO & \makecell*[{{p{2cm}}}]{Maggio\\ et al.~\citep{maggio2024clio}} & Hydra & \makecell*[{{p{3.5cm}}}]{Similar to \citep{jatavallabhula2023conceptfusion} \& scene graph clustering} & \makecell*[{{p{2cm}}}]{Task-dependent submaps} & \makecell*[{{p{2.5cm}}}]{Map generation\\ for task planning} \\

\bottomrule

\end{tabularx}
\caption{Overview and comparison of VLFM applications in semantic mapping by their map representation, the integration of the VLFM, the type of inferences supported, and the application type.}\label{tab:vlfm}
\end{table*}

\subsubsection{NeRF approaches}

\begin{figure}[htb]
\centering
\begin{subfigure}{0.49\textwidth}
    \centering
    \includegraphics[width=0.8\textwidth]{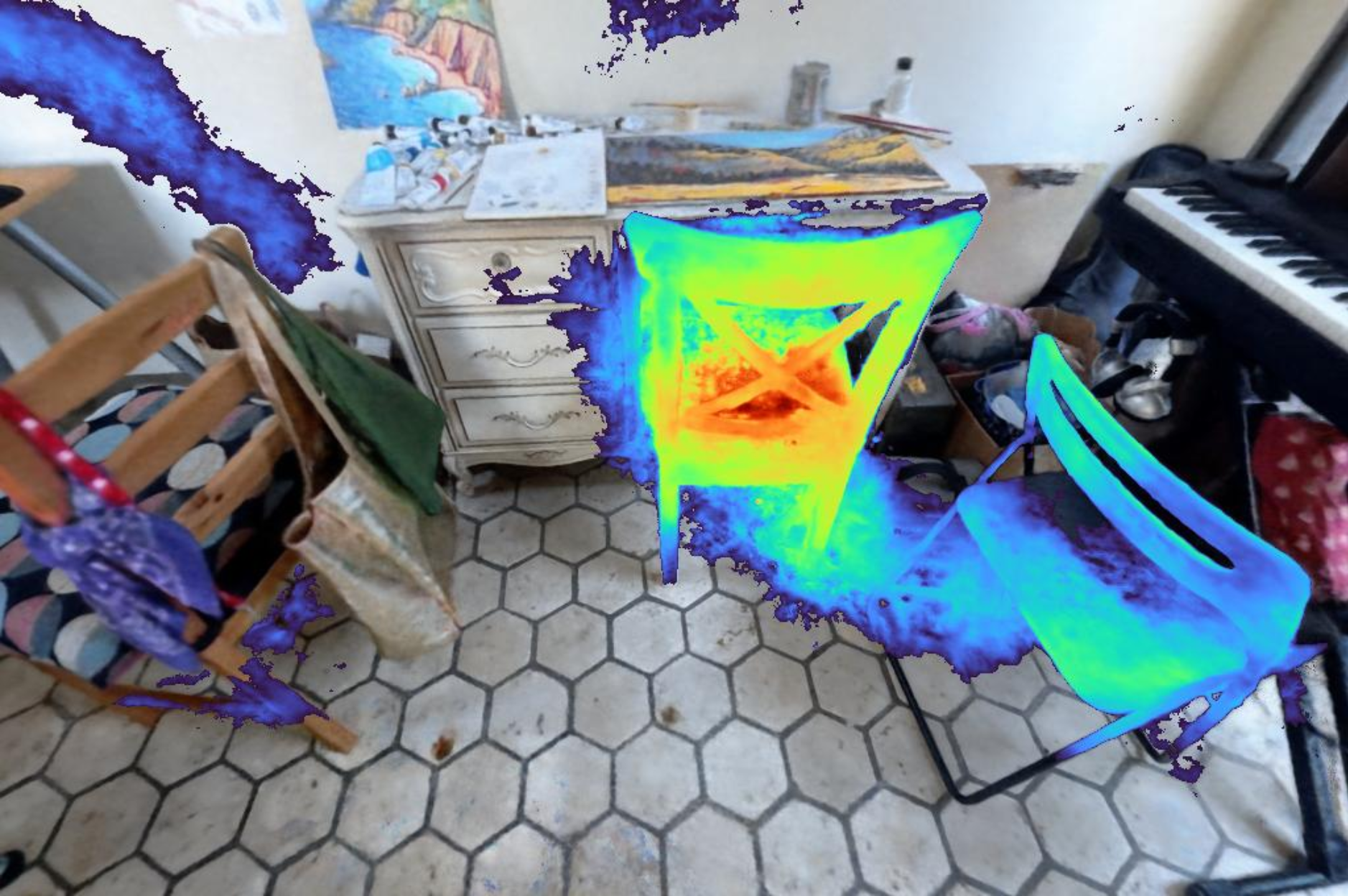}
    \caption{"white chair"}
    \label{fig:LERF:chair}
\end{subfigure}
\begin{subfigure}{0.49\textwidth}
    \centering
    \includegraphics[width=.8\textwidth]{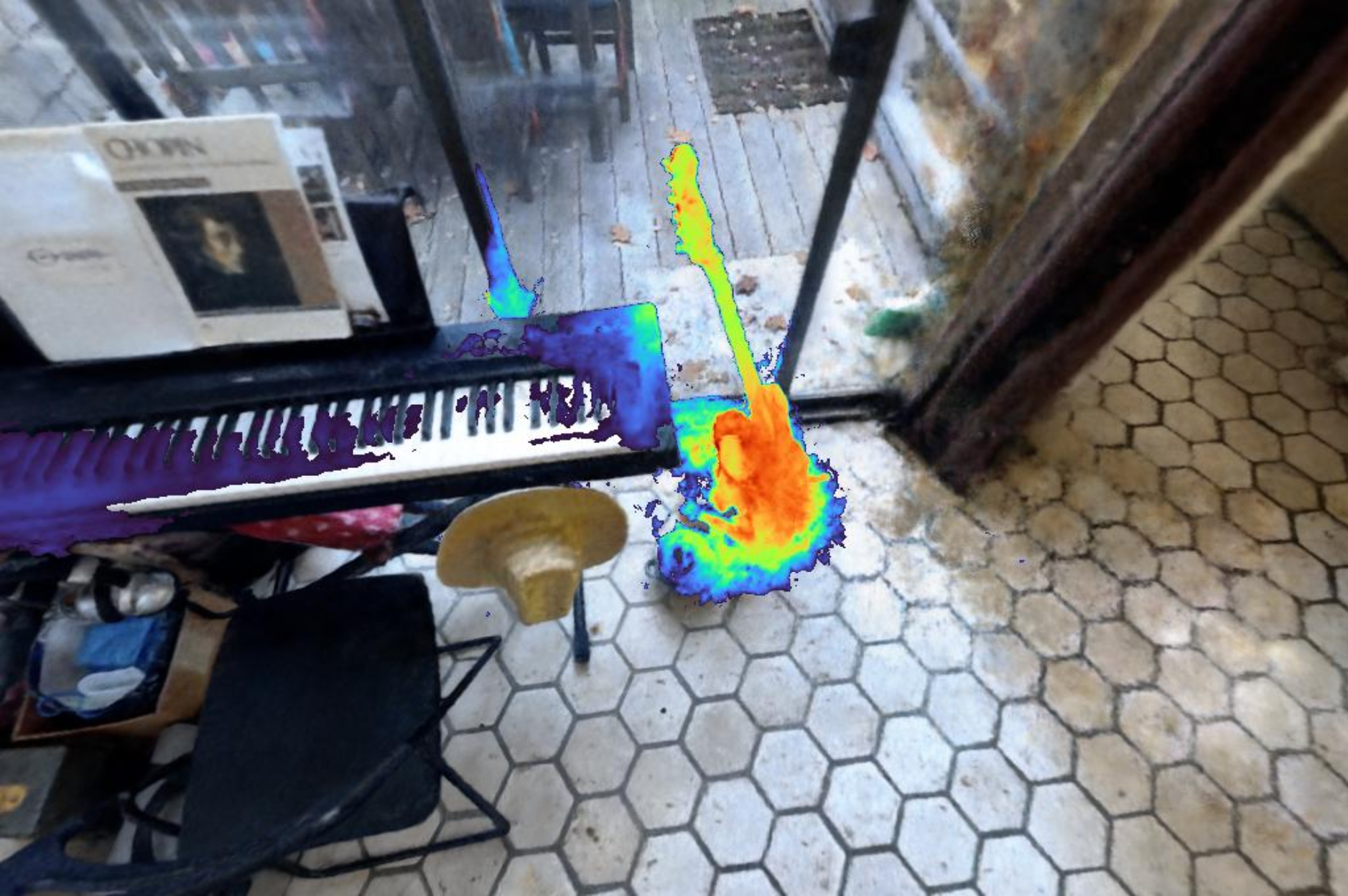}
    \caption{"electric guitar"}
    \label{fig:LERF:guitar}
\end{subfigure}
\caption{Relevancy maps as provided by LERF~\citep{kerrLERFLanguageEmbedded2023} in the "kitchen" dataset from~\citep{tancik2023nerfstudio}.}\label{fig:LERF}
\end{figure}

Approaches based on neural radiance fields (NeRFs) such as LERF~\citep{kerrLERFLanguageEmbedded2023} and CLIP-Fields~\citep{shafiullah2022clip} train models to map CLIP feature vectors to spatial geometry. During training on the images used to reconstruct a 3D scene (\eg from a monocular SLAM system), the models learn to reproduce the geometry from input camera poses and VLFM features. The resulting scene-specific models can be queried with an embedding vector (generated from textual input) to obtain rendered views and saliency maps for the queried objects and locations in the encoded scene (see Fig.~\ref{fig:LERF}).

NeRF-based approaches produce high-quality photo-realistic representations and fine-grained language features that are not limited to the resolution of a 3D sensor. However, these approaches are limited in real-world applications as the geometry and VLFM information are encoded implicitly in the network, requiring extensive training for each scene. They also support neither updates nor fine-tuning of the representation upon \eg changes in the environment or during exploration, and are additionally limited to small scenes and cannot accurately represent larger, more diverse environments.

\subsubsection{Geometry-based approaches}

Geometry-based approaches rely on established 3D reconstruction pipelines to capture the geometry of a scene. Thus, they store the VLFM features directly for each geometric primitive in the map (point, voxel) and therefore do not require training for each scene, allowing them to generalize to many different environments. Additionally, as their map representations can be modified and updated in real time, they are better suited to robotic mapping scenarios. 


OpenScene~\citep{pengOpenScene3DScene2023a} transfers pixel-based features from the 2D domain into the 3D domain by distilling a specialized 3D CNN to reproduce the 2D features on individual 3D points. The resulting aggregated features encode the visual properties of each point in a geometric map representation from multiple views and scales. Similar to LERF, the resulting 3D representation can be queried for saliency maps of arbitrary objects and concepts.

In ConceptFusion~\citep{jatavallabhula2023conceptfusion}, Segment Anything~\citep{kirillov2023segment} or Mask2Former~\citep{cheng2022masked} is used to generate a class-agnostic segmentation of the input images. The resulting bounding boxes are then used to produce crops of the detected objects. Using an off-the-shelf image-level VLFM, a global feature vector for the whole image and local feature vectors for the individual crops are aggregated into pixel-level feature vectors. The resulting segmentation masks are then integrated into a 3D map using a point-based reconstruction pipeline. The resulting map can then be queried for arbitrary objects using multiple modalities if supported by the underlying VLFM. In contrast to OpenScene, the map provides an accurate instance segmentation versus simple saliency maps.

A similar approach is explored in OK-Robot~\cite{liu2024ok}, where the resulting point-based map is utilized in a combined robotic navigation and manipulation task where the robot must find arbitrary objects in a map and move them to new locations. The authors leverage LangSAM \citep{medeirosLucamedeirosLangsegmentanything2024} to accurately segment the object for manipulation once the robot can observe it directly. 

\begin{figure}[htb]
\centering
\includegraphics[width=.4\textwidth]{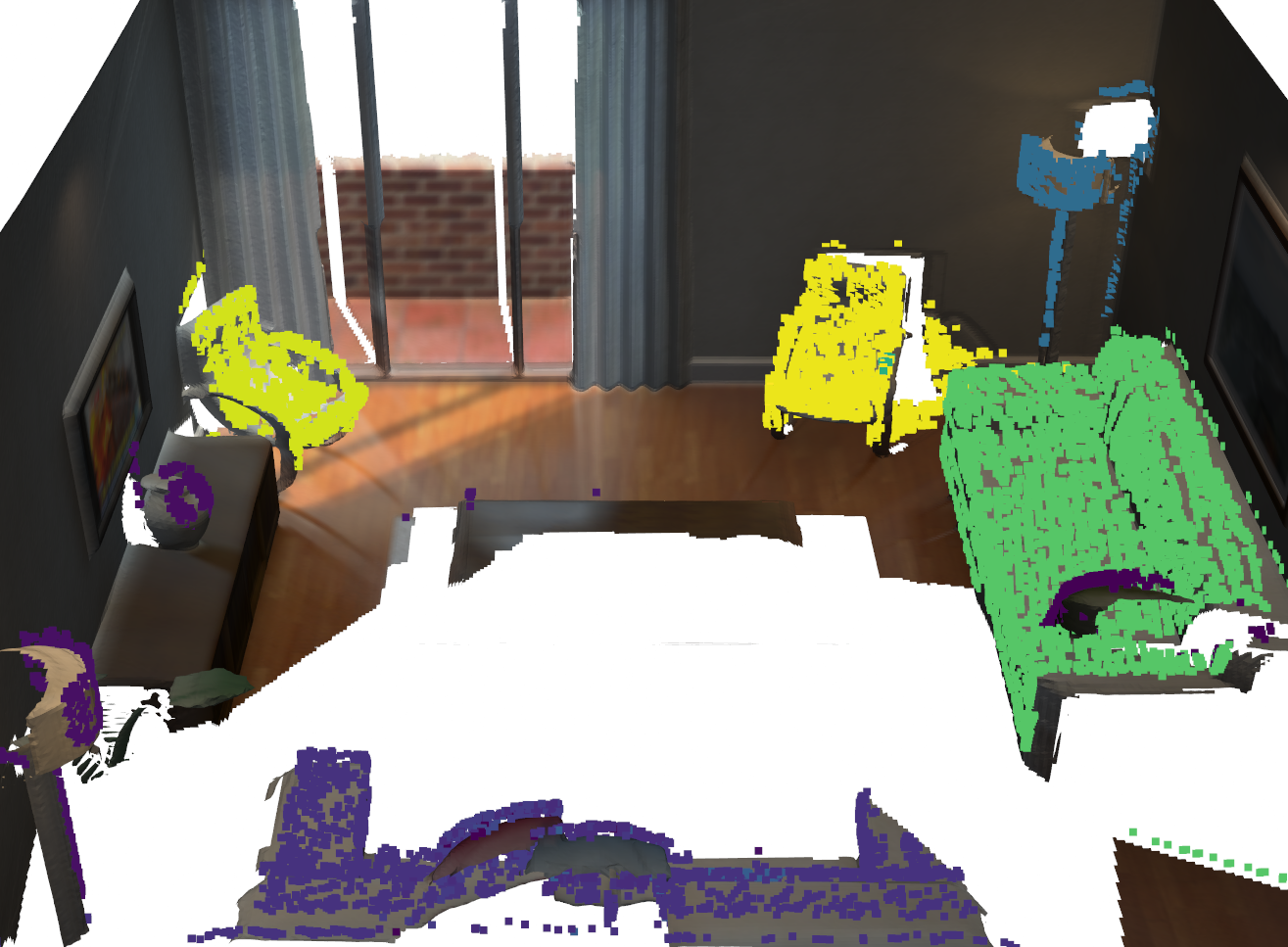}
\caption{Result segments for the query "chair" from a map generated by Open-Fusion~\citep{yamazaki2024open} from the ICL-NUIM dataset~\citep{handa:etal:ICRA2014}. The returned segments are highlighted by their confidence from violet (low) to yellow (high). }\label{fig:open_fusion}
\end{figure}

Open-Fusion~\citep{yamazaki2024open} utilizes the 2D SEEM-Model \citep{zou2024segment} to generate region-based confidence maps from input images. The extracted regions are then used to annotate the voxels of a TSDF-based 3D map. As this region-based segmentation is much faster than offline processing required in previous approaches, Open-Fusion enables the mapping process to be run directly on the 3D sensor data stream. Using region- instead of pixel-based features also means fewer feature vectors must be stored, as only a single vector is needed for each region. An example of a map generated by this approach with a possible query result is shown in Fig.~\ref{fig:open_fusion}.

\subsubsection{Scene-graph approaches}

As geometry-based approaches represent the features of individual points or regions and lack the concept of an object, the resulting maps cannot distinguish individual object instances and only identify groups or parts of objects instead. Changes in the locations and poses of objects are also difficult to represent. To address these issues and support individual object instances, recent approaches leverage semantic scene graphs (see Section~\ref{sec:scene_graphs}) for open-vocabulary 3D maps.

\Citet{werby2024hierarchical} present Hierarchical Open-Vocabulary 3D Scene Graphs (HOV-SG). Their approach constructs a semantic scene graph of a class agnostic segmentation using Hydra \citep{Hughes2022-fg} and stores CLIP features for each segment, similar to \citep{jatavallabhula2023conceptfusion}. The obtained features are aggregated in the higher-level nodes of the scene graph, thus providing a set of representative features for \eg individual rooms and buildings. This approach showed promise in a language-based navigation task where a robot was tasked with finding the location of arbitrary objects in the map given queries in natural language. 


In \citep{guConceptGraphsOpenVocabulary3D2023}, an approach similar to Khronos \citep{Schmid-RSS24-Khronos} is used to construct a scene graph consisting of the point clouds of individual object instances. Each node in the graph is maintained and updated individually, and traditional and class-agnostic segmentation models (\eg SAM) are used to obtain the object segmentation and compute an additional VLFM feature vector from each segment's mask. The scene graph structure is then determined by an LLM (see Section~\ref{subsubsec:llm_applications}).

CLIO~\citep{maggio2024clio} builds upon prior approaches to generate task-dependent semantic maps, focusing on sub-maps of relevant objects versus querying the main map. Sub-maps are generated for a given task by clustering the language features of the objects according to a task specification in natural language. 

\subsubsection{Challenges}

Recent works show that integrating VLFM features into semantic maps is an attractive and powerful approach to exploiting the capabilities of natural language in robotics, and especially to achieve generalization. However, several challenges and shortcomings must be addressed in future research.

As VLFM inference is essentially a neighbor search in an embedding space, it lacks any implicit specialized knowledge about the requested classes. This can lead to unexpected results during inference, especially in uncommon scenarios. 

Additionally, while VLFM-based maps can represent and retrieve arbitrary objects, they are incapable of reasoning except on a very basic level (\eg about materials or color). A semantic map leveraging VLFMs thus still requires additional external knowledge to provide further reasoning capabilities. With the recent advent of Large Multimodal Models, there have been attempts to fuse the strong vision-language representation of VLFMs with the common-sense-like reasoning capabilities that LMMs provide (see Section \ref{subsec:llm}). As of now, the issue of reasoning in VLFMs remains a challenge. 

All the approaches presented here also utilize existing VLFMs, which were trained on massive datasets covering general use cases. Many specialized use cases are thus not supported by these models and require fine-tuning of the model before it can be used in these applications. The performance of VLFMs in these specialized robotics use cases in \eg the agricultural domain is yet to be explored.

\subsection{Large language models}
\label{subsec:llm}





A large language model (LLM) is a computational model that uses statistical methods to analyze and predict the probabilities of word sequences in natural language and which is designed to capture the patterns, grammar, and semantic meaning of natural language \citep{achiam2023gpt}.
Recent models based on the transformer architecture \citep{vaswani2017attention} capture these patterns from massive, internet-scale datasets. Utilizing self-attention, they determine the relationships between preceding tokens, including words, punctuation marks, and sentences, to predict the next output.

Due to their textual focus, interaction with LLMs is typically done in the form of a \textit{chatbot} where the LLMs produce answers to given prompts while still taking former prompts and previous answers into account up to a certain \textit{context size}. Due to the amount of implicit knowledge incidentally encoded in these models during training, LLMs can be applied to many different tasks including text and code generation, summarising existing text, and assisting with decision-making. Though they do not engage in true problem-solving or inference from first principles \citep{kambhampati2024can}, they retrieve information in a manner that mimics reasoning and are effective in assisting robots in retrieving the necessary information to process data and interact with a semantic ontology or a semantic scene graph \citep{rana2023sayplan, ocker2023exploring}.

Large Multimodal Models can process other data modalities beyond text, such as videos and images \citep{yin2023survey, wu2023multimodal}, by integrating other foundation models such as VLFMs. This allows them to reason directly on the kind of multi-modal sensor data typically provided by a robot's perception system. This makes them of particular interest for many robotics tasks including semantic mapping, navigation, and manipulation, and recent years have seen multiple applications using LLMs and LMMs in robotics (summarized in Tab. \ref{tab:lmm}). We outline these early applications in the following sections, focusing on tasks involving semantic mapping. 

\begin{table*}[tbp]
\small
\begin{tabularx}{\textwidth}{p{2cm} p{3cm} p{2cm} X p{2.5cm}}
\toprule

\thead{Name} & \thead{Reference} & \thead{Model} & \thead{Use Case} & \thead{Application} \\

\midrule

- & \makecell*[{{p{3cm}}}]{\Citet{Strader2023-je}} & \makecell*[{{p{2.5cm}}}]{GPT-4~\citep{achiam2023gpt}} & \makecell*[{{p{4.5cm}}}]{Generation of semantic ontologies as training data to classify room types from the contained objects} & \makecell*[{{p{2.5cm}}}]{Large scale 3D scene graph generation\\ (indoor \& outdoor)} \\

\midrule

\makecell*[{{p{2cm}}}]{Explore until Confident} & \makecell*[{{p{2cm}}}]{\Citet{ren2024explore}} & \makecell*[{{p{2.5cm}}}]{Prismatic VLM} & \makecell*[{{p{4.5cm}}}]{Deriving worthy exploration locations from visual data} & \makecell*[{{p{2.5cm}}}]{Embodied question answering through robotic exploration}\\

\midrule

ConceptGraphs & \makecell*[{{p{2cm}}}]{\Citet{guConceptGraphsOpenVocabulary3D2023}} & \makecell*[{{p{2.5cm}}}]{LLaVA~\citep{liu2024visual}} & \makecell*[{{p{4.5cm}}}]{Deriving spatial and semantic relations of objects in the scene} & \makecell*[{{p{2.5cm}}}]{3D semantic mapping and scene graph generation} \\

\midrule

SayPlan & \makecell*[{{p{3cm}}}]{\Citet{rana2023sayplan}} & \makecell*[{{p{2.5cm}}}]{GPT-3.5~\citep{brown2020language},\\ GPT-4~\citep{achiam2023gpt}} & \makecell*[{{p{4.5cm}}}]{Construct relevant semantic scene subgraph} & \makecell*[{{p{2.5cm}}}]{Robotic task plan generation} \\

\midrule

MoMa-LLM & \makecell*[{{p{3cm}}}]{\Citet{honerkamp2024language}} & \makecell*[{{p{2.5cm}}}]{GPT-3.5, GPT-4} & \makecell*[{{p{4.5cm}}}]{Determining next best actions from a semantic scene graph in unknown environments} & \makecell*[{{p{2.5cm}}}]{Robotic exploration} \\

\bottomrule
\end{tabularx}
\caption{Overview and comparison of references using language models to improve semantic mapping and applications.}\label{tab:lmm}
\end{table*}

\subsubsection{Semantic mapping applications}
\label{subsubsec:llm_applications}



LLMs and LMMs have increasingly been utilized in semantic mapping research in recent works. Here, their information retrieval capabilities rather than their capacity for language generation are of primary interest. 

\Citet{Strader2023-je}, for example, 
utilize an LLM in their work as a preprocessing step to infer a semantic ontology for classifying a room's type based on the observed objects that appear in it. 
As building such an ontology manually is time-consuming and error-prone, utilizing the LLM saves a considerable amount of time. The authors query the LLM to choose the $k$ objects in a fixed set of object types most likely to distinguish a certain room type from all others available. This way, they obtain a suitable ontology without extensive manual work or training data. To address the issue of the LLM potentially hallucinating and returning faulty results, the authors filter out invalid answers and repeat a query when necessary. The final obtained ontology is then used in a neuro-symbolic Logic Tensor Network~\citep{Badreddine2020-yd} for training a specialized neural network to perform room classification directly on the sensor data.

\begin{figure}[htb]
\centering
\includegraphics[width=.45\textwidth]{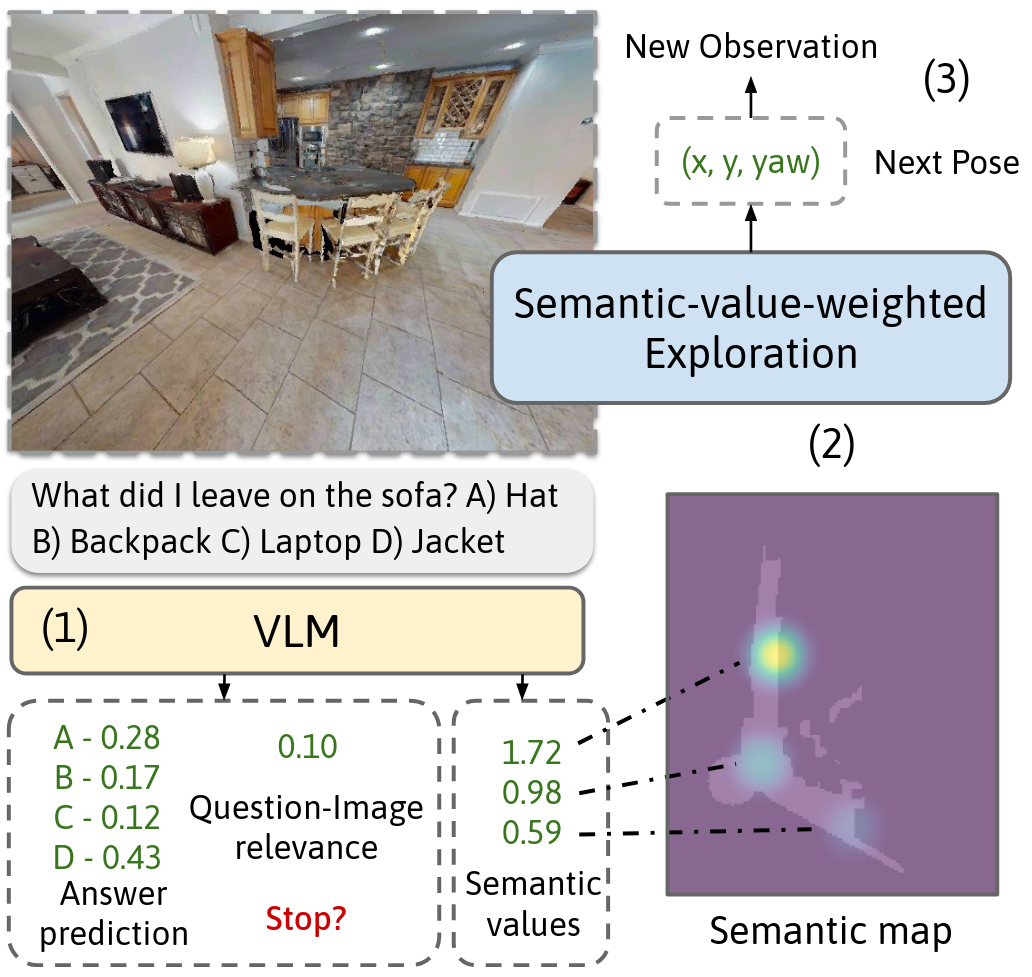}
\caption{Example for Embodied Question Answering using an LLM and a semantic map. Image source: \citep{ren2024explore}}\label{fig:explore_until_confident_example}
\end{figure}

\Citet{ren2024explore} use a multi-modal LLM to perform \textit{embodied question answering} in partially unknown environments. The authors utilize the encoded prior knowledge in the LLM, in conjunction with a semantic map, to generate task-dependent exploration plans, with the tasks given as natural language questions about the environment or objects (see Fig.~\ref{fig:explore_until_confident_example}). In experiments, this approach outperforms both traditional frontier-based and CLIP-enhanced exploration, showing that the utilized LLM is indeed contributing useful knowledge to the task.

ConceptGraphs~\citep{guConceptGraphsOpenVocabulary3D2023} (see also Section~\ref{subsec:scene_graph_prior_knowledge}) use the LMM LLava~\citep{li2024llava} during the semantic mapping process to infer semantic relations between the detected objects using image data, thereby constructing a semantic scene graph. Though the original paper is only a proof of concept with just two available semantic relations (\textit{on} and \textit{below}), the work demonstrates that LMMs are also able to reason about scene structure.

SayPlan~\citep{rana2023sayplan} focuses on plan generation through an LLM-based pipeline that performs semantic searches to construct a task-specific subgraph from a full semantic scene graph. This process reduces the number of input tokens to the LLM, enhancing scalability and reducing cost. An iterative re-planning pipeline using feedback from the reduced scene graph performs multiple queries to the LLM until a valid, executable plan is obtained. Although requiring internet access and therefore not suitable for onboard-only deployment on a robot, the system achieves a maximum accuracy rate of approximately 73\% in complex search tasks in office and home environments.


In a similar work, \citet{honerkamp2024language} propose MoMa-LLM to systematically ground an LLM within an open-vocabulary semantic scene graph constructed dynamically using an approach similar to \citep{werby2024hierarchical}. During operation, the system is tasked with fetching a particular object from an unknown environment, with the LLM being updated at each step with the most recent scene graph and asked for the next suitable action for the robot to perform, ultimately outperforming traditional methods even in real-world environments.

\subsubsection{Challenges}
\label{subsubsec:llm_challenges}

Despite being able to simultaneously consider information from many different sources - a significant advantage over traditional systems - LLMs and LMMs still face several significant issues which presently limit their applicability \citep{ji2023survey, valmeekam2022large, rudin2019stop}.

Firstly, the training and deployment of LLMs require significant computational resources, which are challenging to provide on most robotics platforms. This means that robots utilizing LLMs must have a constant broadband internet connection, limiting their potential, particularly in \eg outdoor scenarios. Additionally, LLMs are prone to hallucinations \citep{huang2023survey}, sometimes causing them to generate inaccurate or unreasonable output that must be filtered out via control mechanisms to prevent unexpected or outright dangerous behavior during a robot's operation. 

In \citep{valmeekam2022large}, it is additionally shown that LLMs do not perform well at reasoning about new problems that are not in their training sets. Recently, Large Reasoning Models (LRMs) have made significant improvements on this point, but at a high computational cost and without guarantees or an extensive capacity to identify when a problem is unsolvable \citep{valmeekam2024llmscantplanlrms}. Further issues related to ethics, safety, data privacy, regulations, cultural bias, and other topics remain critical challenges as well, \cf~\citep{braunschweig2021reflections, helberger2023chatgpt, jiao2024navigating}.

\section{Conclusion}
\label{sec:conclusion}

Knowledge-integrated AI in semantic mapping continues to evolve, moving beyond traditional rule-based systems and towards new applications of knowledge processing to diverse input modalities, a trend evident in the field of semantic mapping. This review provides an overview of two distinct yet complementary aspects of modern semantic mapping, namely semantic scene graphs and language models, and addresses their individual contributions as well as emerging synergies.

Semantic scene graphs enable the structured symbolic representation and processing of scene structure, with the application of GNNs allowing for the efficient integration of even large-scale graph data and prior knowledge into machine learning pipelines, greatly enhancing robotics applications utilizing these scene graphs. VLFM and other multi-modal language models not only bridge the visual and textual domains but also allow the exploitation of the inherent fuzziness and generalizability of natural language in robotic perception, allowing even richer environment representations and enhancing reasoning capabilities. LMMs in turn enable robotic applications to apply common-sense-like reasoning directly on multi-modal input data, alleviating the need for strict, manually created rule bases in robotic applications such as navigation, manipulation, and task planning.

As these fields continue to expand and diversify, we expect significant developments in applications involving multi-modal reasoning, sensor data processing, and natural language interaction, contributing to broader advancement in AI and robotics, particularly with respect to handling the dynamic interplay between symbolic and sub-symbolic approaches.

\section*{Acknowledgments}

This work is supported by the ExPrIS and LIEREx projects through grants from the German Federal Ministry of Education and Research (BMBF) with Grant Numbers 01IW23001 and 01IW24004.
The DFKI Niedersachsen (DFKI NI) is sponsored by the Ministry of Science and Culture of Lower Saxony and the Volkswagen Stiftung.

\bibliographystyle{elsarticle-num-names}
\bibliography{bibliography}

\end{document}